\documentclass[11pt]{article}
\usepackage{coling2016}
\usepackage{times}
\usepackage{url}
\usepackage{latexsym}
\usepackage{graphicx}
\usepackage{listings}
\usepackage{fancyvrb}
\usepackage{wrapfig}
\usepackage{booktabs}

\title{Distant supervision for emotion detection using Facebook reactions}

\author{Chris Pool \\
Anchormen, Groningen\\
The Netherlands\\
  {\tt c.pool@anchormen.nl} \\\And
  Malvina Nissim \\
  CLCG, University of Groningen \\
  The Netherlands \\
  {\tt m.nissim@rug.nl} \\}

\date{}

\begin{document}
\maketitle
\begin{abstract}

We exploit the Facebook reaction feature in a distant supervised fashion to train a support vector machine classifier for emotion detection, using several feature combinations and combining different Facebook pages. We test our models on existing benchmarks for emotion detection and show that \textit{employing only information that is derived completely automatically}, thus without relying on any handcrafted lexicon as it's usually done, we can achieve competitive results. The results also show that there is large room for improvement, especially by gearing the collection of Facebook pages, with a view to the target domain. 

\end{abstract}

\section{Introduction}

\blfootnote{
 
     {\noindent This work is licenced under a Creative Commons 
     Attribution 4.0 International Licence.
     Licence details:
     \url{http://creativecommons.org/licenses/by/4.0/}
    }}

In the spirit of the brevity of social media's messages and reactions, people have got used to express feelings minimally and symbolically, as with hashtags on Twitter and Instagram. On Facebook, people tend to be more wordy, but posts normally receive more simple ``likes'' than longer comments. Since February~2016, Facebook users can express  specific emotions in response to a post thanks to the newly introduced \textit{reaction feature} (see Section~\ref{sec:FBData}), so that now a post can be wordlessly marked with an expression of say ``joy" or ``surprise" rather than a generic ``like''. 

It has been observed that this new feature helps Facebook to know much more about their users and exploit this information for targeted advertising \cite{wired}, but interest in people's opinions and how they feel isn't limited to commercial reasons, as it invests social monitoring, too, including health care and education \cite{SentimentEmotionSurvey2015}. 
However, emotions and opinions are not always expressed this explicitly, so that there is high interest in developing systems towards their automatic detection. Creating manually annotated datasets large enough to train supervised models is not only costly, but also---especially in the case of opinions and emotions---difficult, due to the intrinsic subjectivity of the task \cite{strapparava2008learning,kim2010evaluation}. Therefore, research has focused on unsupervised methods enriched with information derived from lexica, which are manually created \cite{kim2010evaluation,chaffar2011using}. Since \newcite{go2009twitter} have shown that happy and sad emoticons can be successfully used as signals for sentiment labels, \textit{distant supervision}, i.e.  using some reasonably safe signals as proxies for automatically labelling training data \cite{mintz:2009}, has been used also for emotion recognition, for example exploiting both emoticons and Twitter hashtags \cite{purver2012experimenting}, but mainly towards creating emotion lexica. \newcite{mohammad2015using} use hashtags, experimenting also with highly fine-grained emotion sets (up to almost 600 emotion labels), to create the large \textit{Hashtag Emotion Lexicon}. Emoticons are used as proxies also by \newcite{hallsmarmulti}, who use distributed vector representations to find which words are interchangeable with emoticons but also which emoticons are used in a similar context.

We take advantage of distant supervision by using Facebook reactions as proxies for emotion labels, which to the best of our knowledge hasn't been done yet, and we train a set of Support Vector Machine models for emotion recognition. Our models, differently from existing ones, exploit information which is \textit{acquired entirely automatically}, and achieve competitive or even state-of-the-art results for some of the emotion labels on existing, standard evaluation datasets. For explanatory purposes, related work is discussed further and more in detail when we describe the benchmarks for evaluation (Section~\ref{sec:emoData}) and when we compare our models to existing ones (Section~\ref{sec:evaluation}). We also explore and discuss how choosing different sets of Facebook pages as training data provides an intrinsic domain-adaptation method.

\section{Facebook reactions as labels}
\label{sec:FBData}

For years, on Facebook people could leave comments to posts, and also ``like'' them, by using a thumbs-up feature to explicitly express a generic, rather underspecified, approval. A ``like'' could thus mean ``I like what you said", but also ``I like that you bring up such topic (though I find the content of the article you linked annoying)". 
\begin{wrapfigure}{l}{6cm}
  \includegraphics[scale=.2]{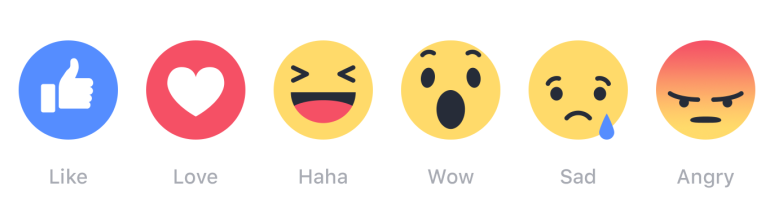}
  \caption{Facebook reactions}
  \label{fig:facebook_reactions}
\end{wrapfigure}
In February 2016, after a short trial, Facebook made a more explicit \textit{reaction} feature available world-wide. Rather than allowing for the underspecified ``like'' as the only wordless response to a post, a set of six more specific reactions was introduced, as shown in Figure \ref{fig:facebook_reactions}: \texttt{Like, Love, Haha, Wow, Sad and Angry}. We use such reactions as proxies for emotion labels associated to posts.

\noindent We collected Facebook posts and their corresponding reactions from public pages using the Facebook API, which we accessed via the Facebook-sdk python library\footnote{https://pypi.python.org/pypi/facebook-sdk}. We chose different pages (and therefore domains and stances), aiming at a balanced and varied dataset, but we did so mainly based on intuition (see Section~\ref{sec:model}) and with an eye to the nature of the datasets available for evaluation (see Section~\ref{sec:evaluation}). The choice of which pages to select posts from is far from trivial, and we believe this is actually an interesting aspect of our approach, as by using different Facebook pages one can intrinsically tackle the domain-adaptation problem (See Section~\ref{sec:conclusions} for further discussion on this). The final collection of Facebook pages for the experiments described in this paper is as follows: 
\texttt{FoxNews}, 
\texttt{CNN}, 
\texttt{ESPN}, 
\texttt{New York Times}, 
\texttt{Time magazine},
\texttt{Huffington Post Weird News}, 
\texttt{The Guardian}, 
\texttt{Cartoon Network}, 
\texttt{Cooking Light}, 
\texttt{Home Cooking Adventure}, 
\texttt{Justin Bieber}, 
\texttt{Nickelodeon}, 
\texttt{Spongebob}, 
\texttt{Disney}.

\begin{wrapfigure}{r}{8cm}
\vspace*{-.8cm}
\begin{Verbatim}[fontsize=\scriptsize]
[
 {
  "created_time": "2016-06-19T01:40:00+0000",
  "message": "Walt Disney World representatives said 
  they plan to put up fencing and signs at all resorts 
  and waterways.",
  "reactions": [5073, 4483, 60, 22, 54, 284, 170, 0]
 }
],
[
 {
  "created_time": "2016-06-19T01:00:00+0000",
  "message": "Charlene and Joseph Handrik face more
  than 550 counts of animal cruelty.",
  "reactions": [2256, 1011, 16, 6, 123, 409, 691, 0]
 }
],
\end{Verbatim}
\vspace*{-.8cm}
\caption{Sample of resulting JSON file\label{fig:json}. The order of values/reactions is \texttt{total, like, love, haha, wow, sad, angry, thankful}.}
\end{wrapfigure}
\footnotetext{Note that \texttt{thankful} was only available during specific time spans related to certain events, as Mother's Day in May 2016.}
\setcounter{footnote}{2}

For each  page, we downloaded the latest 1000 posts, or the maximum available if there are fewer, from February 2016,
retrieving the counts of reactions for each post. The output is a JSON file containing a list of dictionaries with a timestamp, the post and a reaction vector with frequency values, which indicate how many users used that reaction in response to the post (Figure~\ref{fig:json}). The resulting emotion vectors must then be turned into an emotion label.\footnotemark{}

In the context of this experiment, we made the simple decision of associating to each post the emotion with the highest count, ignoring \texttt{like} as it is the default and most generic reaction people tend to use. Therefore, for example, to the first post in Figure~\ref{fig:json}, 
  we would associate the label \texttt{sad}, as it has the highest score (284) among the meaningful emotions we consider, though it also has non-zero scores for other emotions. At this stage, we didn't perform any other entropy-based selection of posts, to be investigated in future work.

\section{Emotion datasets}
\label{sec:emoData}

Three datasets annotated with emotions are commonly used for the development and evaluation of emotion detection systems, namely the \textit{Affective Text} dataset, the \textit{Fairy Tales} dataset, and the \textit{ISEAR} dataset. In order to compare our performance to state-of-the-art results, we have used them as well. In this Section, in addition to a description of each dataset, we provide an overview of the emotions used, their distribution, and how we mapped them to those we obtained from Facebook posts in Section~\ref{sec:overview}. A summary is provided in Table~\ref{overview_data-sets}, which also shows, in the bottom row, what role each dataset has in our experiments: apart from the development portion of the Affective Text, which we used to develop our models (Section~\ref{sec:model}), all three have been used as benchmarks for our evaluation.

\subsection{Affective Text dataset}
\label{sec:data:affect}
Task~14 at SemEval~2007  \cite{strapparava2007semeval} was concerned with the classification of emotions and valence in news headlines. The headlines where collected from several news websites including Google news,  The New York Times, BBC News and CNN. The used emotion labels were \texttt{Anger, Disgust, Fear, Joy, Sadness, Surprise}, in line with the six basic emotions of Ekman's standard model \cite{ekman1992argument}. Valence was to be determined as positive or negative. Classification of emotion and valence were treated as separate tasks. 
Emotion labels were not considered as mututally exclusive, and each emotion was assigned a score from 0~to~100. Training/developing data amounted to 250 annotated headlines (\textit{Affective development}), while systems were evaluated on another 1000 (\textit{Affective test}). Evaluation was done using two different methods: a fine-grained evaluation using Pearson's \textit{r}  to measure the correlation between the system scores and the gold standard; and a coarse-grained method where each emotion score was converted to a binary label, and precision, recall, and f-score were computed to assess performance. As it is done in most works that use this dataset \cite{kim2010evaluation,chaffar2011using,calvo2013emotions}, we also treat this as a classification problem (coarse-grained).  This dataset has been extensively used for the evaluation of various unsupervised methods \cite{strapparava2008learning}, but also for testing different supervised learning techniques and feature portability \cite{mohammad:2012:NAACL-HLT}.

\subsection{Fairy Tales dataset}
This is a dataset collected by \newcite{alm2008affect}, where about 1,000 sentences from fairy tales (by B. Potter, H.C. Andersen and Grimm) were annotated with the same six emotions of the Affective Text dataset, though with different names: \texttt{Angry}, \texttt{Disgusted}, \texttt{Fearful}, \texttt{Happy}, \texttt{Sad}, and \texttt{Surprised}. In most works that use this dataset \cite{kim2010evaluation,chaffar2011using,calvo2013emotions}, only sentences where all annotators agreed are used, and the labels \texttt{angry} and \texttt{disgusted} are merged. We adopt the same choices.

\subsection{ISEAR}
\label{sec:data:isear}
The ISEAR (International Survey on Emotion Antecedents and Reactions \cite{scherer1994evidence,scherer1997role}) is a dataset created in the context of a psychology project of the 1990s, by collecting questionnaires answered by people with different cultural backgrounds.   The main aim of this project was to gather insights in cross-cultural aspects of emotional reactions. Student respondents, both psychologists and non-psychologists, were asked to report situations in which they had experienced all of seven major emotions (\texttt{joy, fear, anger, sadness, disgust, shame} and \texttt {guilt}). In each case, the questions covered the way they had appraised a given situation and how they reacted. The final dataset contains reports by approximately 3000 respondents from all over the world, 
for a total of 7665 sentences labelled with an emotion, making this the largest dataset out of the three we use.

\subsection{Overview of datasets and emotions}
\label{sec:overview}

We summarise datasets and emotion distribution from two viewpoints. First, because there are different sets of emotions labels in the datasets and Facebook data, we need to provide a mapping and derive a subset of emotions that we are going to use for the experiments. 
This is shown in Table~\ref{overview_data-sets}, where in the ``Mapped'' column we report the final emotions we use in this paper: \texttt{anger, joy, sadness, surprise}. All labels in each dataset are mapped to these final emotions, which are therefore the labels we use for training and testing our models. 

\begin{figure}[hbt]
\begin{minipage}{.43\textwidth}
  \includegraphics[scale=.40]{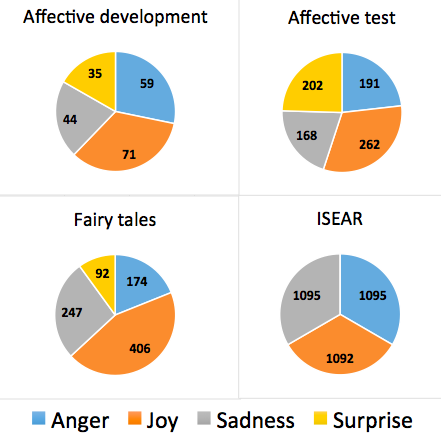}
  \caption{Emotion distribution in the datasets  \label{fig:distribution_data-sets}}
\end{minipage}
~
\begin{minipage}{.53\textwidth}
    \includegraphics[scale=.365,keepaspectratio]{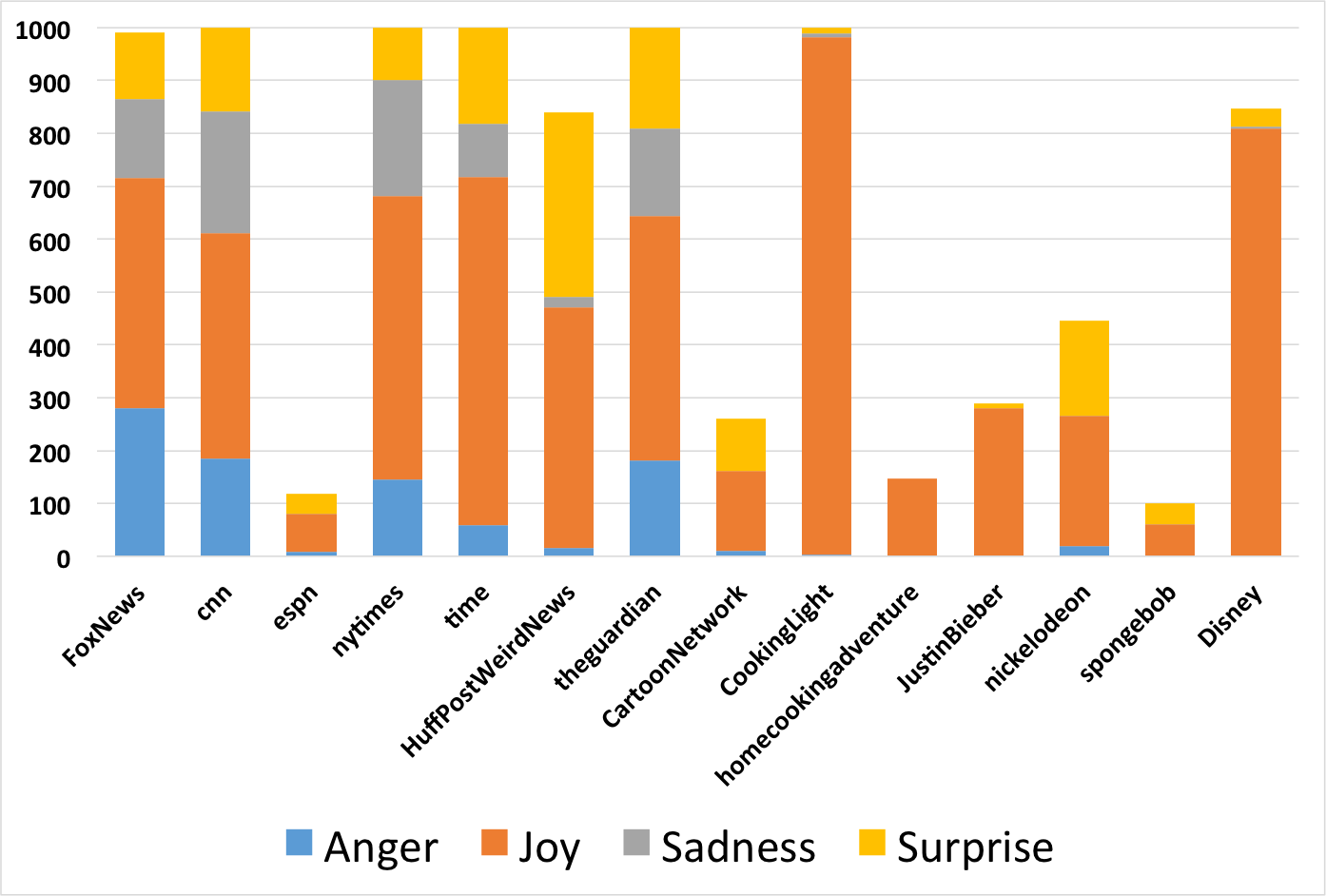}  
\caption{Emotion distribution per Facebook page\label{fig:distribution_facebook}}
\end{minipage}
\end{figure}

Second, the distribution of the emotions for each dataset is different, as can be seen in Figure~\ref{fig:distribution_data-sets}. 

\begin{wraptable}{r}{12.6cm}
\vspace*{-.5cm}
\caption{Emotion labels in existing datasets, Facebook, and resulting mapping for the experiments in this work. The last row indicates which role each dataset has in our experiments.\label{overview_data-sets}}
\centering
\begin{tabular}{lllll}
\toprule
\textbf{Affective Text} & \textbf{Fairy tales} & \textbf{ISEAR} & \textbf{Facebook} & \textbf{Mapped} \\ \hline
Anger            & Angry-Disgusted      & Anger               & Angry   & \texttt{anger}          \\ \hline
Disgust          & Angry-Disgusted      & Disgust               &    &  \texttt{anger}              \\ \hline
Fear             & Fearful              & Fear               &      &              \\ \hline
Joy              & Happy                & Joy                & Haha, Love & \texttt{joy}\\ \hline
Sadness          & Sad                  & Sadness               & Sad     & \texttt{sadness}          \\ \hline
Surprise         & Suprised             &                & Wow          & \texttt{surprise}     \\ 
\hline
                 &                      & Shame               &       &         \\ \hline
                 &                      & Guilt               &       &         \\                  
\bottomrule
\textbf{development/test} &          \textbf{test}                             & \textbf{test}                    & \textbf{train}  & \\ \bottomrule
\end{tabular}
\end{wraptable}
\noindent In Figure~\ref{fig:distribution_facebook} we also provide the distribution of the emotions \texttt{anger, joy, sadness, surprise} per Facebook page, in terms of number of posts (recall that we assign to a post the label corresponding to the majority emotion associated to it, see Section~\ref{sec:FBData}). We can observe that for example pages about news  tend to have more sadness and anger posts, while pages about cooking and tv-shows have a high percentage of joy posts. We will use this information to find the best set of pages for a given target domain (see Section~\ref{sec:evaluation}).

\vspace*{-.5cm}

\section{Model}
\label{sec:model}

There are two main decisions to be taken in developing our model: (i) which Facebook pages to select as training data, and (ii) which features to use to train the model, which we discuss below. Specifically, we first set on a subset of pages and then experiment with features. Further exploration of the interaction between choice of pages and choice of features is left to future work, and partly discussed in Section~\ref{sec:conclusions}. For development, we use a small portion of the Affective data set described in Section~\ref{sec:data:affect}, that is the portion that had been released as development set for SemEval's 2007 Task~14 \cite{strapparava2007semeval}, which contains 250 annotated sentences (\textit{Affective development}, Section~\ref{sec:data:affect}). All results reported in this section are on this dataset.
  The test set of Task~14 as well as the other two datasets described in Section~\ref{sec:emoData} will be used to evaluate the final models (Section~\ref{sec:model}).

\subsection{Selecting Facebook pages}
\label{sec:selecting}

Although page selection is a crucial ingredient of this approach, which we believe calls for further and deeper, dedicated investigation, for the experiments described here we took a rather simple approach. First, we selected the pages that would provide training data based on intuition and availability, then chose different combinations according to results of a basic model run on development data, and eventually tested feature combinations, still on the development set.

For the sake of simplicity and transparency, we first trained an SVM with a simple bag-of-words model and default parameters as per the Scikit-learn implementation \cite{scikit-learn} on different combinations of pages. Based on results of the attempted combinations as well as on the distribution of emotions in the development dataset (Figure~\ref{fig:distribution_data-sets}), we selected a \textit{best model} (\textbf{B-M}), namely the combined set of \texttt{Time, The Guardian} and \texttt{Disney}, which yields the highest results on development data. \texttt{Time} and \texttt{The Guardian} perform well on most emotions but \texttt{Disney} helps to boost the performance for the \texttt{Joy} class.

\subsection{Features}
In selecting appropriate features, we mainly relied on previous work and intuition. We experimented with different combinations, and all tests were still done on \textit{Affective development}, using the pages for the best model (\textbf{B-M}) described above as training data. 
Results are in Table~\ref{test_results_svm}.
Future work will further explore the simultaneous selection of features and page combinations.

\paragraph{Standard textual features} We use a set of basic text-based features to capture the emotion class. These include a tf-idf bag-of-words feature, word (2-3) and character (2-5) ngrams, and features related to the presence of negation words, and to the usage of punctuation.

\paragraph{Affect Lexicons} This feature is used in all unsupervised models as a source of information, and we mainly include it to assess its contribution, but eventually do not use it in our final model. 
 
We used the NRC10 Lexicon  because it performed best in the experiments by \cite{mohammad:2012:NAACL-HLT}, which is built around the emotions \texttt{anger}, \texttt{anticipation}, \texttt{disgust}, \texttt{fear}, \texttt{joy}, \texttt{sadness}, and \texttt{surprise}, and the valence values \texttt{positive} and \texttt{negative}. For each word in the lexicon, a boolean value indicating presence or absence is associated to each emotion. For a whole sentence, a global score per emotion can be obtained by summing the vectors for all content words of that sentence included in the lexicon, and used as feature.

\paragraph{Word Embeddings}
As additional feature, we also included Word Embeddings, namely distributed representations of words in a vector space, which have been exceptionally successful in boosting performance in a plethora of NLP tasks.
We use three different embeddings: 

\begin{itemize} 

\item \textit{Google embeddings}: pre-trained embeddings trained on Google~News and obtained with the skip-gram architecture described in  \cite{mikolov2013distributed}. This model contains 300-dimensional vectors for 3 million words and phrases.

\item \textit{Facebook embeddings}: embeddings that we trained on our scraped Facebook pages for a total of 20,000 sentences. Using the \texttt{gensim} library \cite{gensim}, we trained the embeddings with the following parameters: window size of 5, learning rate of 0.01 and dimensionality of 100. We filtered out words with frequency lower than 2 occurrences.

\item \textit{Retrofitted embeddings}: 
Retrofitting \cite{retrofitting} has been shown as a simple but efficient way of informing trained embeddings with additional information derived from some lexical resource, rather than including it directly at the training stage, as it's done for example to create sense-aware \cite{iacobacci2015sensembed} or sentiment-aware \cite{tang:14} embeddings.\footnote{Training emotion-aware embeddings is a strategy that we plan to explore in future work.} In this work, we retrofit general embeddings to include information about emotions, so that emotion-similar words can get closer in space. Both the Google as well as our Facebook embeddings were retrofitted with lexical information obtained from the NRC10~Lexicon mentioned above, which provides emotion-similarity for each token. Note that differently from the previous two types of embeddings, the retrofitted ones do rely on handcrafted information in the form of a lexical resource.

\end{itemize}

\subsection{Results on development set}

We report precision, recall, and f-score on the development set. The average f-score is reported as \textit{micro-average}, to better account for the skewed distribution of the classes as well as in accordance to what is usually reported for this task \cite{mohammad2015using}.

\begin{table*}[!htbp]
\caption{Results on the development set  (\textit{Affective development}). \textit{avg f} is the micro-averaged f-score.\label{test_results_svm}}
\centering
\begin{tabular}{l|c|c|c|c|c}
\toprule
\hline
    & \texttt{\textbf{anger}} & \texttt{\textbf{joy}} & \texttt{\textbf{sadness}} & \texttt{\textbf{surprise}} &  \\ 
    \cline{2-5}
          \textbf{Feature}         & prec,rec,f & prec,rec,f & prec,rec,f & prec,rec,f & avg f \\

\hline
  \footnotesize{Tf-idf} &
  \footnotesize{0.57,0.22,0.32}  & 
  \footnotesize{0.44,0.51,0.47} & 
  \footnotesize{0.41,0.25, 0.31} & 
  \footnotesize{0.22,0.49,0.30} & 
  \footnotesize{{0.368}}  \\

\hline
  \footnotesize{Lexicon} &
  \footnotesize{0.28,0.08,0.13}  & 
  \footnotesize{0.43,0.37,0.40} & 
  \footnotesize{0.31,0.30, 0.30} & 
  \footnotesize{0.20,0.51,0.29} & 
   \footnotesize{{0.297}}  \\ 

\hline
  \footnotesize{Token n-grams(2,5)} &
  \footnotesize{0.00,0.00,0.00}  & 
  \footnotesize{1.00,0.01,0.03} & 
  \footnotesize{0.00,0.00, 0.00} & 
  \footnotesize{0.17,1.00,0.29} & 
  \footnotesize{{0.172}}  \\

\hline
  \footnotesize{Character n-grams(2,5)} &
  \footnotesize{0.50,0.03,0.06}  & 
  \footnotesize{0.39,0.73,0.51} & 
  \footnotesize{0.38,0.07, 0.12} & 
  \footnotesize{0.17,0.31,0.22} & 
  \footnotesize{{0.325}}  \\

\hline
  \footnotesize{All features} &
  \footnotesize{0.40,0.03,0.06}  & 
  \footnotesize{0.35,0.97,0.52} & 
  \footnotesize{0.62,0.11, 0.19} & 
  \footnotesize{1.00,0.03,0.06} & 
  \footnotesize{{0.368}}  \\

\midrule
\midrule

  \footnotesize{Google (G) embeddings} &
  \footnotesize{0.41,0.49,0.45}  & 
  \footnotesize{0.56,0.46,0.51} & 
  \footnotesize{0.48,0.57, 0.52} & 
  \footnotesize{0.22,0.17,0.19} & 
  \footnotesize{{0.445}}\\

\hline
  \footnotesize{Facebook (FB) embeddings} &
  \footnotesize{0.33,0.15,0.21}  & 
  \footnotesize{0.31,0.45,0.37} & 
  \footnotesize{0.23,0.11,0.15} & 
  \footnotesize{0.20,0.31,0.24} & 
  \footnotesize{{0.273}} \\

\hline
  \footnotesize{Retrofitted G-embeddings} &
  \footnotesize{0.36,0.20,0.26}  & 
  \footnotesize{0.42,0.48,0.45} & 
  \footnotesize{0.30,0.25, 0.27} & 
  \footnotesize{0.20,0.34,0.26} & 
  \footnotesize{{0.330}}\\

\hline
  \footnotesize{Retrofitted FB-embeddings} &
  \footnotesize{0.07,0.02,0.03}  & 
  \footnotesize{0.34,0.86,0.49} & 
  \footnotesize{0.36,0.09,0.15} & 
  \footnotesize{0.17,0.03,0.05} & 
  \footnotesize{{0.321}} \\ 

\midrule
\midrule

  \footnotesize{Tf-idf + G-emb} &
  \footnotesize{0.42,0.46,0.44}  & 
  \footnotesize{0.45,0.49,0.47} & 
  \footnotesize{0.49,0.41, 0.44} & 
  \footnotesize{0.29,0.26,0.27} & 
  \footnotesize{{0.426}}  \\

\hline
  \footnotesize{All features + G-emb} &
  \footnotesize{0.63,0.29,0.40}  & 
  \footnotesize{0.43,0.83,0.56} & 
  \footnotesize{0.46,0.27, 0.34} & 
  \footnotesize{0.33,0.17,0.23} & 
  \footnotesize{{0.450}}  \\

\hline
  \footnotesize{All features -- Lexicon + G-emb} &
  \footnotesize{0.62,0.34,0.44}  & 
  \footnotesize{0.43,0.85,0.57} & 
  \footnotesize{0.57,0.30, 0.39} & 
  \footnotesize{0.36,0.14,0.20} & 
  \footnotesize{\textbf{0.469}}  \\

\hline
\bottomrule           
\end{tabular}
\end{table*}

\noindent From Table~\ref{test_results_svm} we draw three main observations. First, a simple tf-idf bag-of-word mode works already very well, to the point that  the other textual and lexicon-based features don't  seem to contribute to the overall f-score (0.368), although there is a rather substantial variation of scores per class. Second, Google embeddings perform a lot better than Facebook embeddings, and this is likely due to the size of the corpus used for training. Retrofitting doesn't seem to help at all for the Google embeddings, but it does boost the Facebook embeddings, leading to think that with little data, more accurate task-related information is helping, but corpus size matters most. Third, in combination with embeddings, all features work better than just using tf-idf, but removing the Lexicon feature, which is the only one based on hand-crafted resources, yields even better results. Then our best model (\textbf{B-M}) on development data relies \textit{entirely on automatically obtained information}, both in terms of training data as well as features.

\section{Results}
\label{sec:evaluation}
In Table~\ref{final_results} we report the results of our model on the three datasets standardly used for the evaluation of emotion classification, which we have described in Section~\ref{sec:emoData}.

Our \textbf{B-M} model relies on subsets of Facebook pages for training, which were chosen according to their performance on the development set as well as on the observation of emotions distribution on different pages and in the different datasets, as described in Section~\ref{sec:model}. The feature set we use is our best on the development set, namely 
all the features plus Google-based embeddings, but excluding the lexicon. This makes our approach completely independent of any manual annotation or handcrafted resource. 
Our model's performance is compared to the following systems, for which results are reported in the referred literature. Please note that no other existing model was re-implemented, and results are those  reported in the respective papers.

\paragraph{\newcite{kim2010evaluation}} experiment with four different unsupervised techniques that rely on lexicon-derived information. In Table~\ref{final_results} we report the scores for their best average performing approach, namely a CNMF-based categorical classification. They made the decision not to deal with \texttt{surprise} because this emotion is not present in the ISEAR dataset.

\paragraph{\newcite{strapparava2008learning}} experiment with several models based on a core LSA model and, in their best performing model (\texttt{LSA-all emotion words}) whose results we report in Table~\ref{final_results}, also use information from lexical resources both in their general (WordNet \cite{wordnet}) and emotion-aware (WordNet~Affect \cite{strapparava2004wordnet}) form.

\paragraph{\newcite{danisman2008feeler}} adopt a supervised approach, training a model using the ISEAR dataset and testing it on the Affective text dataset. They only report results per category in terms of f-score, without further specification of how precision and recall contribute.

\bigskip

\noindent We have mentioned that the selection of Facebook pages is relevant and can be also thought of as a tool for domain adaptation in accordance with the characteristics of the target domains/datasets (see also Section~\ref{sec:FBData} and Figures~\ref{fig:distribution_data-sets}--\ref{fig:distribution_facebook}). Although we believe that such an interesting aspect will require deeper investigation (see also Section~\ref{sec:conclusions}), we preliminary test this assumption by developing and comparing two more models: a model that uses a combination of pages that we expect will perform best on the Fairy Tales dataset (\textbf{FT-M}), and a model that uses a combination of pages that should perform best on the ISEAR dataset (\textbf{ISE-M}). The feature set is kept the same for all three models.

\paragraph{FT-M}
The sentences in the Fairy Tales dataset are quite different compared to the news headlines in the development set. Looking at the distribution in this dataset, as can be seen in Figure~\ref{fig:distribution_data-sets}, \texttt{Joy} is the most frequent class. We selected the pages \texttt{HuffPostWeirdNews, ESPN} and \texttt{CNN} for this model especially looking at the performance for the emotions that are most frequent in this dataset.

\paragraph{ISE-M}
As described in Section~\ref{sec:data:isear}, the sentences in the ISEAR collection are also different compared to the two other datasets. Looking at the distribution in Figure~\ref{fig:distribution_data-sets} and according to performance on relevant emotions (we took into account the absence of \texttt{Surprise} in this dataset), we selected the pages \texttt{Time, The Guardian} and \texttt{CookingLight} for this model.

\begin{table}[!h]
\caption{Results on test datasets according to \textbf{P}recision, \textbf{R}ecall and \textbf{F}-score.\label{final_results}}
\centering
\begin{tabular}{|l||l|l|l||l|l|l||l|l|l|}
\cline{2-10}
    \multicolumn{1}{}{{}} & 
    \multicolumn{3}{|c||}{\textbf{Affective test}} & 
    \multicolumn{3}{|c||}{\textbf{Fairy Tales}} & 
    \multicolumn{3}{|c|}{\textbf{ISEAR}}  \\
        
\cline{2-10}
\cline{2-10}                      
     \multicolumn{1}{c|}{{}} & 
    \textbf{P} & 
    \textbf{R} & 
    \textbf{F} & 
    \textbf{P} & 
    \textbf{R} & 
    \textbf{F} & 
    \textbf{P} & 
    \textbf{R} & 
    \textbf{F} \\ 
     
\hline                     

&    \multicolumn{9}{|c|}{\texttt{\textbf{anger}}} \\                 

                                        \hline
    \tiny{B-M} & 
    \footnotesize{0.50} & 
    \footnotesize{0.35} & 
    \footnotesize{\textbf{0.41}} & 
    \footnotesize{0.33} & 
    \footnotesize{0.04} & 
    \footnotesize{0.07} & 
    \footnotesize{0.72} & 
    \footnotesize{0.06} & 
    \footnotesize{0.11} \\ 

\hline
    \tiny{FT-M} & 
    \footnotesize{\textbf{0.51}} & 
    \footnotesize{0.30} & 
    \footnotesize{0.38} &
    \footnotesize{0.27} & 
    \footnotesize{0.02} & 
    \footnotesize{0.04} & 
    \footnotesize{0.57} & 
    \footnotesize{0.10} & 
    \footnotesize{0.17} \\ 

\hline
    \tiny{ISE-M} & 
    \footnotesize{0.48} & 
    \footnotesize{0.35} & 
    \footnotesize{0.40} & 
    \footnotesize{0.36} & 
    \footnotesize{0.05} & 
    \footnotesize{0.08} & 
    \footnotesize{\textbf{0.74}} & 
    \footnotesize{0.06} & 
    \footnotesize{0.11} \\

\hline
    \tiny{\cite{strapparava2008learning} } &
    \footnotesize{0.06} & 
    \footnotesize{\textbf{0.88}} & 
    \footnotesize{0.12} &
    &
    &
    &
    &
    &
    \\

\hline
    \tiny{\cite{kim2010evaluation} } &
    \footnotesize{0.29} & 
    \footnotesize{0.26} & 
    \footnotesize{{0.28}} &
    \footnotesize{\textbf{0.77}} & 
    \footnotesize{\textbf{0.56}} & 
    \footnotesize{\textbf{0.65}} &
    \footnotesize{0.41} & 
    \footnotesize{\textbf{0.99}} & 
    \footnotesize{\textbf{0.58}} \\
    
\hline
    \tiny{\cite{danisman2008feeler} } &
    \footnotesize{} & 
    \footnotesize{} & 
    \footnotesize{0.24} &
    \footnotesize{} & 
    \footnotesize{} & 
    \footnotesize{} &
    \footnotesize{} & 
    \footnotesize{} & 
    \footnotesize{} \\

\hline
\hline

 &   \multicolumn{9}{|c|}{\texttt{\textbf{joy}}} \\ 

\hline
    \tiny{B-M} & 
    \footnotesize{0.39} & 
    \footnotesize{0.85} & 
    \footnotesize{0.54} & 
    \footnotesize{0.49} & 
    \footnotesize{{0.77}} & 
    \footnotesize{0.60} & 
    \footnotesize{{0.41}} & 
    \footnotesize{{0.79}} & 
    \footnotesize{{0.53}} \\ 

\hline
    \tiny{FT-M} & 
    \footnotesize{0.41} & 
    \footnotesize{0.77} & 
    \footnotesize{0.54} & 
    \footnotesize{0.49} & 
    \footnotesize{0.69} & 
    \footnotesize{0.58} &
    \footnotesize{\textbf{0.42}} & 
    \footnotesize{0.63} & 
    \footnotesize{0.50} \\ 

\hline
    \tiny{ISE-M} & 
    \footnotesize{0.39} & 
    \footnotesize{0.82} & 
    \footnotesize{0.53} & 
    \footnotesize{0.48} & 
    \footnotesize{\textbf{0.81}} & 
    \footnotesize{0.60} & 
    \footnotesize{{0.40}} & 
    \footnotesize{\textbf{0.83}} & 
    \footnotesize{\textbf{0.54}} \\ 

\hline 
    \tiny{\cite{strapparava2008learning} } &
    \footnotesize{0.19} & 
    \footnotesize{\textbf{0.90}} & 
    \footnotesize{0.31} &
    &
    &
    &
    &
    &
    \\

\hline
    \tiny{\cite{kim2010evaluation}} &
    \footnotesize{\textbf{0.77}} & 
    \footnotesize{0.58} & 
    \footnotesize{\textbf{0.65}} &
    \footnotesize{\textbf{0.80}} & 
    \footnotesize{{0.76}} & 
    \footnotesize{\textbf{0.78}} &
    \footnotesize{0.39} & 
    \footnotesize{0.01} & 
    \footnotesize{0.01} \\
    
\hline
    \tiny{\cite{danisman2008feeler}} &
    \footnotesize{} & 
    \footnotesize{} & 
    \footnotesize{0.50} &
    \footnotesize{} & 
    \footnotesize{} & 
    \footnotesize{} &
    \footnotesize{} & 
    \footnotesize{} & 
    \footnotesize{} \\
    
\hline
\hline
&    \multicolumn{9}{|c|}{\texttt{\textbf{sadness}}} \\ 
    
\hline
    \tiny{B-M} & 
    \footnotesize{0.51} & 
    \footnotesize{0.21} & 
    \footnotesize{0.30} & 
    \footnotesize{0.43} & 
    \footnotesize{0.39} & 
    \footnotesize{0.41} & 
    \footnotesize{0.50} & 
    \footnotesize{\textbf{0.39}} & 
    \footnotesize{\textbf{0.44}} \\ 

\hline
    \tiny{FT-M} & 
    \footnotesize{\textbf{0.53}} & 
    \footnotesize{0.28} & 
    \footnotesize{0.37} & 
    \footnotesize{0.50} & 
    \footnotesize{0.24} & 
    \footnotesize{0.33} & 
    \footnotesize{\textbf{0.79}} & 
    \footnotesize{0.28} & 
    \footnotesize{0.41} \\ 

\hline
    \tiny{ISE-M} & 
    \footnotesize{0.49} & 
    \footnotesize{0.21} & 
    \footnotesize{0.29} & 
    \footnotesize{0.43} & 
    \footnotesize{0.34} & 
    \footnotesize{0.38} & 
    \footnotesize{0.51} & 
    \footnotesize{0.38} & 
    \footnotesize{\textbf{0.44}} \\ 

\hline
    \tiny{\cite{strapparava2008learning} } &
    \footnotesize{0.12} & 
    \footnotesize{\textbf{0.87}} & 
    \footnotesize{0.22} &
    &
    &
    &
    &
    &
    \\

\hline
    \tiny{\cite{kim2010evaluation} } &
    \footnotesize{0.50} & 
    \footnotesize{0.45} & 
    \footnotesize{\textbf{0.48}} &
    \footnotesize{\textbf{0.71}} & 
    \footnotesize{\textbf{0.82}} & 
    \footnotesize{\textbf{0.77}} &
    \footnotesize{0.37} & 
    \footnotesize{0.01} & 
    \footnotesize{0.25} \\

\hline
    \tiny{\cite{danisman2008feeler}} &
    \footnotesize{} & 
    \footnotesize{} & 
    \footnotesize{0.37} &
    \footnotesize{} & 
    \footnotesize{} & 
    \footnotesize{} &
    \footnotesize{} & 
    \footnotesize{} & 
    \footnotesize{} \\

\hline
\hline
&    \multicolumn{9}{|c|}{\texttt{\textbf{surprise}}} \\                               
    
\hline
  \tiny{B-M} & 
  \footnotesize{0.20} & 
  \footnotesize{0.05} & 
  \footnotesize{0.08} & 
  \footnotesize{0.12} & 
  \footnotesize{0.04} & 
  \footnotesize{0.06} & 
  \footnotesize{} & 
  \footnotesize{} & 
  \footnotesize{} \\ 

\hline
    \tiny{FT-M} & 
    \footnotesize{0.25} & 
    \footnotesize{0.17} & 
    \footnotesize{\textbf{0.20}} & 
    \footnotesize{\textbf{0.14}} & 
    \footnotesize{\textbf{0.33}} & 
    \footnotesize{\textbf{0.19}} & 
    \footnotesize{} & 
    \footnotesize{} & 
    \footnotesize{} \\ 

\hline
    \tiny{ISE-M} & 
    \footnotesize{\textbf{0.27}} & 
    \footnotesize{0.08} & 
    \footnotesize{0.12} & 
    \footnotesize{0.17} & 
    \footnotesize{0.04} & 
    \footnotesize{0.07} & 
    \footnotesize{} & 
    \footnotesize{} & 
    \footnotesize{} \\

\hline
    \tiny{\cite{strapparava2008learning} } &
    \footnotesize{0.08} & 
    \footnotesize{\textbf{0.95}} & 
    \footnotesize{0.14} &
    &
    &
    &
    &
    &
    \\

\hline
    \tiny{\cite{kim2010evaluation} } &
    & 
    & 
    &
    &
    &
    &
    &
    &
    \\
    
\hline
    \tiny{\cite{danisman2008feeler} } &
    \footnotesize{} & 
    \footnotesize{} & 
    \footnotesize{} &
    \footnotesize{} & 
    \footnotesize{} & 
    \footnotesize{} &
    \footnotesize{} & 
    \footnotesize{} & 
    \footnotesize{} \\

\hline
\hline

 &   \multicolumn{9}{|c|}{{AVERAGE (micro f-score)}} \\                               
    
\hline
  \tiny{B-M} & 
  \multicolumn{3}{c|}{  \footnotesize{0.409} } &
  \multicolumn{3}{c|}{  \footnotesize{0.459} } &
  \multicolumn{3}{c|}{  \footnotesize{0.411} } 
\\ 

\hline
    \tiny{FT-M} & 
      \multicolumn{3}{|c|}{  \footnotesize{\textbf{0.412}} } &
  \multicolumn{3}{|c|}{  \footnotesize{0.408} } &
  \multicolumn{3}{|c|}{  \footnotesize{0.336} } 
\\

\hline
    \tiny{ISE-M} & 
      \multicolumn{3}{|c|}{  \footnotesize{0.405} } &
  \multicolumn{3}{|c|}{  \footnotesize{\textbf{0.460}} } &
  \multicolumn{3}{c|}{  \footnotesize{\textbf{0.422}} } 
\\ 

\hline

\end{tabular}
\end{table}

\bigskip

\noindent 
In Table~\ref{final_results} we report results for all of the models mentioned above. We indicate averages only for our models, since not all approaches deal with the same sets of emotions and we cannot easily compute them. 
We discuss results both in terms of how our models fair with respect to other systems as reported the literature, as well as how they compare to one another with a view to the selection of Facebook pages.

Compared to other systems, our models are globally competitive, given that \textbf{B-M} is entirely unsupervised. Overall, the unsupervised but heavily lexicon-based best model of \cite{kim2010evaluation} performs well on all emotions, excluding surprise, which they do not address (thus also making their classification task slightly easier). Differently from existing systems,  our models appear rather balanced in terms of performance on the different emotions as well as in precision and recall, and are able to deal well with the variance of the datasets. 

On the Affective Text dataset, we have the highest precision for all emotions but \texttt{joy}, though on this emotion our models have very good recall. The highest recall for all emotions for this dataset is reported in \cite{strapparava2008learning}, together with extremely low precision. Such skewed performance for all emotions can only be explained if different emotion-specific models were trained rather than a single multiclass model, but this is not described as such in the paper. The authors state that their models are completely unsupervised, which is true in terms of training data, but they nevertheless augment them with information derived from hand-crafted resources.

On the Fairy Tales dataset, \cite{kim2010evaluation} 
\newcite{chaffar2011using} also used the Fairy tales dataset to evaluate a supervised model using features like bag-of-words, N-grams and lexical emotion features, but report cross-validated results using accuracy only, and are therefore harder to compare.

On the ISEAR dataset, which is the largest, our models perform best for all emotions but \texttt{anger}, for which however we achieve the highest precision with all our models.

\noindent From the perspective of comparing our models, we do not observe any real correlation between our actual best performances and the models designed to best perform on a given dataset. For example, \textbf{B-M} was expected to perform best on the Affective Text, but it is outperformed by \textbf{FT-M} in the precision of detecting anger and sadness, and overall for the detection of surprise. Generally, by looking at averages, it seems that our best performing model across datasets is \textbf{ISE-M}. However, the extremely large variance among scores for the same emotion on the three datasets, highlights the differences among such datasets  and the need to better tailor training data to different domains. The large discrepancy in detecting different emotions in the same dataset also deserves further investigation. We discuss such issues further in the next section, with a view to future work.

\section{Discussion, conclusions and future work}
\label{sec:conclusions}
We have explored the potential of using Facebook reactions in a distant supervised setting to perform emotion classification. The evaluation on standard benchmarks shows that models trained as such, especially when enhanced with continuous vector representations, can achieve competitive results without relying on any handcrafted resource. An interesting aspect of our approach is the view to domain adaptation via the selection of Facebook pages to be used as training data.

We believe that this approach has a lot of potential, and we see the following directions for improvement. Feature-wise, we want to train emotion-aware embeddings, in the vein of work by \newcite{tang:14}, and \newcite{iacobacci2015sensembed}. Retrofitting FB-embeddings trained on a larger corpus might also be successful, but would rely on an external lexicon. 

The largest room for yielding not only better results but also interesting insights on extensions of this approach lies in the choice of training instances, both in terms of Facebook pages to get posts from, as well as in which posts to select from the given pages. 
For the latter, one could for example only select posts that have a certain length, ignore posts that are only quotes or captions to images, or expand posts by including content from linked html pages, which might provide larger and better contexts \cite{plank:2014}. Additionally, and most importantly, one could use an entropy-based measure to select only posts that have a strong emotion rather than just considering the majority emotion as training label. For the former, namely the choice of Facebook pages, which we believe deserves the most investigation, one could explore several avenues, especially in relation to \textit{stance}-based issues \cite{stance}. In our dataset, for example, a post about Chile beating Colombia in a football match during the Copa America had very contradictory reactions, depending on which side readers would cheer for. Similarly, the very same political event, for example, would get very different reactions from readers if it was posted on Fox News or The Late Night Show, as the target audience is likely to feel very differently about the same issue. This also brings up theoretical issues related more generally to the definition of the emotion detection task, as it's strongly dependent on personal traits of the audience.
Also, in this work, pages initially selected on availability and intuition were further grouped into sets to make training data according to performance on development data, and label distribution. Another criterion to be exploited would be \textit{vocabulary overlap} between the pages and the datasets.

Lastly, we could develop single models for each emotion, treating the problem as a multi-label task. This would even better reflect the ambiguity  and subjectivity intrinsic to assigning emotions to text, where content could be at same time joyful or sad, depending on the reader.

\bigskip

\section*{Acknowledgements}
In addition to the anonymous reviewers, we want to thank Lucia Passaro and Barbara Plank for insightful discussions, and for providing comments on draft versions of this paper.

\bibliographystyle{acl}
\bibliography{bibliography}

\begin{thebibliography}{}

\bibitem[\protect\citename{Alm}2008]{alm2008affect}
Ebba Cecilia~Ovesdotter Alm.
\newblock 2008.
\newblock {\em Affect in text and speech}.
\newblock ProQuest.

\bibitem[\protect\citename{Calvo and Mac~Kim}2013]{calvo2013emotions}
Rafael~A Calvo and Sunghwan Mac~Kim.
\newblock 2013.
\newblock Emotions in text: dimensional and categorical models.
\newblock {\em Computational Intelligence}, 29(3):527--543.

\bibitem[\protect\citename{Chaffar and Inkpen}2011]{chaffar2011using}
Soumaya Chaffar and Diana Inkpen.
\newblock 2011.
\newblock Using a heterogeneous dataset for emotion analysis in text.
\newblock In {\em Advances in Artificial Intelligence}, pages 62--67. Springer.

\bibitem[\protect\citename{Danisman and Alpkocak}2008]{danisman2008feeler}
Taner Danisman and Adil Alpkocak.
\newblock 2008.
\newblock Feeler: Emotion classification of text using vector space model.
\newblock In {\em AISB 2008 Convention Communication, Interaction and Social
  Intelligence}, volume~1, page~53.

\bibitem[\protect\citename{Ekman}1992]{ekman1992argument}
Paul Ekman.
\newblock 1992.
\newblock An argument for basic emotions.
\newblock {\em Cognition \& emotion}, 6(3-4):169--200.

\bibitem[\protect\citename{Faruqui \bgroup et al.\egroup }2015]{retrofitting}
Manaal Faruqui, Jesse Dodge, Sujay~Kumar Jauhar, Chris Dyer, Eduard Hovy, and
  Noah~A. Smith.
\newblock 2015.
\newblock Retrofitting word vectors to semantic lexicons.
\newblock In {\em Proceedings of the 2015 Conference of the North American
  Chapter of the Association for Computational Linguistics: Human Language
  Technologies}, pages 1606--1615, Denver, Colorado, May--June. Association for
  Computational Linguistics.

\bibitem[\protect\citename{Fellbaum}1998]{wordnet}
Christiane Fellbaum.
\newblock 1998.
\newblock {\em WordNet}.
\newblock Wiley Online Library.

\bibitem[\protect\citename{Go \bgroup et al.\egroup }2009]{go2009twitter}
Alec Go, Richa Bhayani, and Lei Huang.
\newblock 2009.
\newblock Twitter sentiment classification using distant supervision.
\newblock {\em CS224N Project Report, Stanford}, 1:12.

\bibitem[\protect\citename{Hallsmar and Palm}2016]{hallsmarmulti}
Fredrik Hallsmar and Jonas Palm.
\newblock 2016.
\newblock Multi-class sentiment classification on twitter using an emoji
  training heuristic.
\newblock Technical report, KTH/Skolan f\"{o}r datavetenskap och kommunikation
  (CSC).
\newblock University essay.

\bibitem[\protect\citename{Iacobacci \bgroup et al.\egroup
  }2015]{iacobacci2015sensembed}
Ignacio Iacobacci, Mohammad~Taher Pilehvar, and Roberto Navigli.
\newblock 2015.
\newblock Sensembed: learning sense embeddings for word and relational
  similarity.
\newblock In {\em Proceedings of ACL}, pages 95--105.

\bibitem[\protect\citename{Kim \bgroup et al.\egroup }2010]{kim2010evaluation}
Sunghwan~Mac Kim, Alessandro Valitutti, and Rafael~A Calvo.
\newblock 2010.
\newblock Evaluation of unsupervised emotion models to textual affect
  recognition.
\newblock In {\em Proceedings of the NAACL HLT 2010 Workshop on Computational
  Approaches to Analysis and Generation of Emotion in Text}, pages 62--70.
  Association for Computational Linguistics.

\bibitem[\protect\citename{Mikolov \bgroup et al.\egroup
  }2013]{mikolov2013distributed}
Tomas Mikolov, Ilya Sutskever, Kai Chen, Greg~S Corrado, and Jeff Dean.
\newblock 2013.
\newblock Distributed representations of words and phrases and their
  compositionality.
\newblock In {\em Advances in neural information processing systems}, pages
  3111--3119.

\bibitem[\protect\citename{Mintz \bgroup et al.\egroup }2009]{mintz:2009}
Mike Mintz, Steven Bills, Rion Snow, and Dan Jurafsky.
\newblock 2009.
\newblock Distant supervision for relation extraction without labeled data.
\newblock In {\em Proceedings of the Joint Conference of the 47th Annual
  Meeting of the ACL and the 4th International Joint Conference on Natural
  Language Processing of the AFNLP: Volume 2 - Volume 2}, ACL '09, pages
  1003--1011, Stroudsburg, PA, USA. Association for Computational Linguistics.

\bibitem[\protect\citename{Mohammad and Kiritchenko}2015]{mohammad2015using}
Saif~M Mohammad and Svetlana Kiritchenko.
\newblock 2015.
\newblock Using hashtags to capture fine emotion categories from tweets.
\newblock {\em Computational Intelligence}, 31(2):301--326.

\bibitem[\protect\citename{Mohammad \bgroup et al.\egroup }2016]{stance}
Saif~M. Mohammad, Svetlana Kiritchenko, Parinaz Sobhani, Xiaodan Zhu, and Colin
  Cherry.
\newblock 2016.
\newblock A dataset for detecting stance in tweets.
\newblock In {\em Proceedings of 10th edition of the the Language Resources and
  Evaluation Conference (LREC)}, Portoro\v{z}, Slovenia.

\bibitem[\protect\citename{Mohammad}2012]{mohammad:2012:NAACL-HLT}
Saif Mohammad.
\newblock 2012.
\newblock Portable features for classifying emotional text.
\newblock In {\em Proceedings of the 2012 Conference of the North American
  Chapter of the Association for Computational Linguistics: Human Language
  Technologies}, pages 587--591, Montr\'{e}al, Canada, June. Association for
  Computational Linguistics.

\bibitem[\protect\citename{Mohammad}2016]{SentimentEmotionSurvey2015}
Saif~M. Mohammad.
\newblock 2016.
\newblock Sentiment analysis: Detecting valence, emotions, and other affectual
  states from text.
\newblock In Herb Meiselman, editor, {\em Emotion Measurement}. Elsevier.

\bibitem[\protect\citename{Pedregosa \bgroup et al.\egroup }2011]{scikit-learn}
F.~Pedregosa, G.~Varoquaux, A.~Gramfort, V.~Michel, B.~Thirion, O.~Grisel,
  M.~Blondel, P.~Prettenhofer, R.~Weiss, V.~Dubourg, J.~Vanderplas, A.~Passos,
  D.~Cournapeau, M.~Brucher, M.~Perrot, and E.~Duchesnay.
\newblock 2011.
\newblock Scikit-learn: Machine learning in {P}ython.
\newblock {\em Journal of Machine Learning Research}, 12:2825--2830.

\bibitem[\protect\citename{Plank \bgroup et al.\egroup }2014]{plank:2014}
Barbara Plank, Dirk Hovy, Ryan~T McDonald, and Anders S{\o}gaard.
\newblock 2014.
\newblock Adapting taggers to twitter with not-so-distant supervision.
\newblock In {\em COLING}, pages 1783--1792.

\bibitem[\protect\citename{Purver and Battersby}2012]{purver2012experimenting}
Matthew Purver and Stuart Battersby.
\newblock 2012.
\newblock Experimenting with distant supervision for emotion classification.
\newblock In {\em Proceedings of the 13th Conference of the European Chapter of
  the Association for Computational Linguistics}, pages 482--491. Association
  for Computational Linguistics.

\bibitem[\protect\citename{{\v R}eh{\r u}{\v r}ek and Sojka}2010]{gensim}
Radim {\v R}eh{\r u}{\v r}ek and Petr Sojka.
\newblock 2010.
\newblock {Software Framework for Topic Modelling with Large Corpora}.
\newblock In {\em {Proceedings of the LREC 2010 Workshop on New Challenges for
  NLP Frameworks}}, pages 45--50, Valletta, Malta, May. ELRA.
\newblock \url{http://is.muni.cz/publication/884893/en}.

\bibitem[\protect\citename{Scherer and Wallbott}1994]{scherer1994evidence}
Klaus~R Scherer and Harald~G Wallbott.
\newblock 1994.
\newblock Evidence for universality and cultural variation of differential
  emotion response patterning.
\newblock {\em Journal of personality and social psychology}, 66(2):310.

\bibitem[\protect\citename{Scherer}1997]{scherer1997role}
Klaus~R Scherer.
\newblock 1997.
\newblock The role of culture in emotion-antecedent appraisal.
\newblock {\em Journal of personality and social psychology}, 73(5):902.

\bibitem[\protect\citename{Stinson}2016]{wired}
Liz Stinson.
\newblock 2016.
\newblock Facebook reactions, the totally redesigned like button, is here.
\newblock {\em Wired}.
\newblock
  \url{http://www.wired.com/2016/02/facebook-reactions-totally-redesigned-like-button/}.

\bibitem[\protect\citename{Strapparava and
  Mihalcea}2007]{strapparava2007semeval}
Carlo Strapparava and Rada Mihalcea.
\newblock 2007.
\newblock Semeval-2007 task 14: Affective text.
\newblock In {\em Proceedings of the 4th International Workshop on Semantic
  Evaluations}, pages 70--74. Association for Computational Linguistics.

\bibitem[\protect\citename{Strapparava and
  Mihalcea}2008]{strapparava2008learning}
Carlo Strapparava and Rada Mihalcea.
\newblock 2008.
\newblock Learning to identify emotions in text.
\newblock In {\em Proceedings of the 2008 ACM symposium on Applied computing},
  pages 1556--1560. ACM.

\bibitem[\protect\citename{Strapparava \bgroup et al.\egroup
  }2004]{strapparava2004wordnet}
Carlo Strapparava, Alessandro Valitutti, et~al.
\newblock 2004.
\newblock Wordnet affect: an affective extension of wordnet.
\newblock In {\em LREC}, volume~4, pages 1083--1086.

\bibitem[\protect\citename{Tang \bgroup et al.\egroup }2014]{tang:14}
Duyu Tang, Furu Wei, Nan Yang, Ming Zhou, Ting Liu, and Bing Qin.
\newblock 2014.
\newblock Learning sentiment-specific word embedding for {T}witter sentiment
  classification.
\newblock In {\em Proceedings of the 52nd Annual Meeting of the Association for
  Computational Linguistics}, volume~1, pages 1555--1565.

\end{thebibliography}

\end{document}